\documentclass{article}

\usepackage{microtype}
\usepackage{graphicx}
\usepackage{subcaption}
\usepackage{booktabs} 
\usepackage{amssymb}
\usepackage{amsmath}
\usepackage{tikz}
\usetikzlibrary{shapes.misc, positioning}
\usepackage{wrapfig,lipsum,booktabs}
\usepackage{arydshln}
\usepackage{hyperref}

\newcommand{\maximize}{\operatorname{maximize}}
\newcommand{\minimize}{\operatorname{minimize}}

\newcommand{\E}{\mathbb{E}}

\newcommand\independent{\protect\mathpalette{\protect\independenT}{\perp}}
\def\independenT#1#2{\mathrel{\rlap{$#1#2$}\mkern2mu{#1#2}}}

\usepackage[accepted]{icml2018}

\icmltitlerunning{Learning Adversarially Fair and Transferable Representations}

\begin{document}

\twocolumn[
\icmltitle{Learning Adversarially Fair and Transferable Representations}



\icmlsetsymbol{equal}{*}

\begin{icmlauthorlist}
\icmlauthor{David Madras$^*$}{ut,vec}
\icmlauthor{Elliot Creager$^*$}{ut,vec}
\icmlauthor{Toniann Pitassi}{ut,vec}
\icmlauthor{Richard Zemel}{ut,vec}
\end{icmlauthorlist}

\icmlaffiliation{ut}{Department of Computer Science, University of Toronto, Toronto, Canada}
\icmlaffiliation{vec}{Vector Institute, Toronto, Canada}

\icmlcorrespondingauthor{David Madras}{madras@cs.toronto.edu}

\icmlkeywords{fairness, representation learning, adversarial learning}

\vskip 0.3in
]



\printAffiliationsAndNotice{\icmlEqualContribution} 

\begin{abstract}
In this paper, we advocate for representation learning as the key to mitigating unfair prediction outcomes downstream.
Motivated by a scenario where learned representations are used by third parties with unknown objectives, we propose and explore adversarial representation learning as a natural method of ensuring those parties act fairly.
We connect group fairness (demographic parity, equalized odds, and equal opportunity) to different adversarial objectives.
Through worst-case theoretical guarantees and experimental validation, we show that the choice of this objective is crucial to fair prediction.
Furthermore, we present the first in-depth experimental demonstration of fair transfer learning and demonstrate empirically that our learned representations admit fair predictions on new tasks while maintaining utility, an essential goal of fair representation learning.
\end{abstract}

\section{Introduction} \label{intro}
There are two implicit steps involved in every prediction task: acquiring data in a suitable form, and specifying an algorithm that learns to predict well given the data.
In practice these two responsibilities are often assumed by distinct parties.
For example, in online advertising the so-called \emph{prediction vendor} profits by selling its predictions (e.g., person $X$ is likely to be interested in product $Y$) to an advertiser, while the \emph{data owner} profits by selling a predictively useful dataset to the prediction vendor \cite{dwork2012fairness}. 

Because the prediction vendor seeks to maximize predictive accuracy, it may (intentionally or otherwise) bias the predictions to unfairly favor certain groups or individuals.
The use of machine learning in this context is especially concerning because of its reliance on historical datasets that include patterns of previous discrimination and societal bias.
Thus there has been a flurry of recent work from the machine learning community focused on defining and quantifying these biases and proposing new prediction systems that mitigate their impact.

Meanwhile, the data owner also faces a decision that critically affects the predictions: what is the correct \emph{representation} of the data?
Often, this choice of representation is made at the level of data collection: feature selection and 
measurement. 
If we want to maximize the prediction vendor's utility, then the right choice is to simply collect and provide the prediction vendor with as much data as possible. 
However, assuring that prediction vendors learn only \textit{fair} predictors complicates the data owner's choice of representation, which must yield predictors that are never unfair but nevertheless have relatively high utility.

In this paper, we frame the data owner's choice as a representation learning problem with an adversary criticizing potentially unfair solutions.
Our contributions are as follows:
We connect common group fairness metrics (demographic parity, equalize odds, and equal opportunity) to adversarial learning by providing appropriate adversarial objective functions for each metric that upper bounds the unfairness of arbitrary downstream classifiers in the limit of adversarial training;
we distinguish our algorithm from previous approaches to adversarial fairness and discuss its suitability to fair classification due to the novel choice of adversarial objective and emphasis on representation as the focus of adversarial criticism;
we validate experimentally that classifiers trained naively (without fairness constraints) from representations learned by our algorithm achieve their respective fairness desiderata;
furthermore, we show empirically that these representations achieve \textit{fair transfer} --- they admit fair predictors on unseen tasks, even when those predictors are not explicitly specified to be fair. 

In Sections \ref{sec:background} and \ref{related-work} we discuss relevant background materials and related work.
In Section \ref{model} we describe our model and motivate our learning algorithm.
In Section \ref{theory} we discuss our novel adversarial objective functions, connecting them to common group fairness metrics and providing theoretical guarantees.
In Section \ref{experiments} we discuss experiments demonstrating our method's success in fair classification and fair transfer learning.

\section{Background} \label{sec:background}
\subsection{Fairness} \label{fair-background}

In fair classification we have some data $X \in \mathbb{R}^n$, labels $Y \in \{0, 1\}$, and sensitive attributes $A \in \{0, 1\}$. 
The predictor outputs a prediction $\hat{Y} \in \{0, 1\}$. 
We seek to learn to predict outcomes that are accurate with respect to $Y$ but fair with respect to $A$; that is, the predictions are accurate but not biased in favor of one group or the other.

There are many possible criteria for group fairness in this context.
One is \textit{demographic parity}, which ensures that the positive outcome is given to the two groups at the same rate, i.e. $P(\hat{Y} = 1 | A = 0) = P(\hat{Y} = 1 | A = 1)$. 
However, the usefulness of demographic parity can be limited if the \textit{base rates} of the two groups differ, i.e. if $P(Y = 1| A = 0) \neq  P(Y = 1| A = 1)$. 
In this case, we can pose an alternate criterion by conditioning the metric on the ground truth $Y$, yielding \textit{equalized odds} and \textit{equal opportunity} \citep{hardt2016equality}; the former requires equal false positive and false negative rates between the groups while the latter requires only one of these equalities. Equal opportunity is intended to match errors in the ``advantaged'' outcome across groups; whereas \citet{hardt2016equality} chose $Y=1$ as the advantaged outcome, the choice is domain specific and we here use $Y=0$ instead without loss of generality.
Formally, this is $P(\hat{Y} \neq Y | A = 0, Y = y) = P(\hat{Y} \neq Y | A = 1, Y = y)\ \forall \ y \in \{0, 1\}$ (or just $y = 0$ for equal opportunity).

Satisfying these constraints is known to conflict with learning well-calibrated classifiers \citep{chouldechova2017fair,kleinberg2016inherent,pleiss2017fairness}.
It is common to instead optimize a relaxed objective \citep{kamishima2012fairness}, whose hyperparameter values negotiate a tradeoff between maximizing utility (usually classification accuracy) and fairness.

\subsection{Adversarial Learning} \label{adv-background}

Adversarial learning is a popular method of training neural network-based models. 
\citet{goodfellow2014generative} framed learning a deep generative model as a two-player game between a generator $G$ and a discriminator $D$. 
Given a dataset $X$, the generator aims to fool the discriminator by generating convincing synthetic data, i.e., starting from random noise $z \sim p(z)$, $G(z)$ resembles $X$. 
Meanwhile, the discriminator aims to distinguish between real and synthetic data by assigning $D(G(z)) = 0$ and $D(X) = 1$.
Learning proceeds by the max-min optimization of the joint objective
\begin{equation}\nonumber
    V(D, G) \triangleq \mathbb{E}_{p(X)} [\log(D(X))] + \mathbb{E}_{p(z)}  [\log(1 - D(G(z)))]
,
\end{equation}
where $D$ and $G$ seek to maximize and minimize this quantity, respectively.

\section{Related Work} \label{related-work}
Interest in fair machine learning is burgeoning as researchers seek to define and mitigate unintended harm in automated decision making systems.
Definitional works have been broadly concerned with \textit{group fairness} or \textit{individual fairness}.
\citet{dwork2012fairness} discussed individual fairness within the owner-vendor framework we utilize. 
\citet{zemel2013learning} encouraged elements of both group and individual fairness via a regularized objective.
An intriguing body of recent work unifies the individual-group dichotomy by exploring fairness at the intersection of multiple group identities, and among small subgroups of individuals \cite{kearns2017preventing,hebert2017calibration}.

\citet{calmon2017optimized} and \citet{hajian2015discrimination} explored fair machine learning by pre- and post-processing training datasets.
\citet{mcnamara2017provably} provides a framework where the data producer, user, and regulator have separate concerns, and discuss fairness properties of representations. 
\citet{louizos2015variational} give a method for learning fair representations with deep generative models by using maximum mean discrepancy \citep{gretton2007kernel} to eliminate disparities between the two sensitive groups.

Adversarial training for deep generative modeling was popularized by \citet{goodfellow2014generative} and applied to deep semi-supervised learning \citep{salimans2016improved,odena2016semi} and segmentation \citep{luc2016semantic}, although similar concepts had previously been proposed for unsupervised and supervised learning \citep{schmidhuber1992learning,gutmann2010noise}.
\citet{ganin2016domain} proposed adversarial representation learning for domain adaptation, which resembles fair representation learning in the sense that multiple distinct data distributions (e.g., demographic groups) must be expressively modeled by a single representation.

\citet{edwards2015censoring} made this connection explicit by proposing adversarially learning a classifier that achieves demographic parity. 
This work is the most closely related to ours, and we discuss some key differences in sections \ref{sec:compare-storkey}.
Recent work has explored the use of adversarial training to other notions of group fairness.
\citet{beutel2017data} explored the particular fairness levels achieved by the algorithm from \citet{edwards2015censoring}, and demonstrated that they can vary as a function of the demographic unbalance of the training data.
In work concurrent to ours, \citet{zhang2018mitigating} use an adversary which attempts to predict the sensitive variable solely based on the classifier output, to learn an equal opportunity fair classifier.
Whereas they focus on fairness in classification outcomes, in our work we allow the adversary to work directly with the learned representation, which we show yields fair and transferable representations that in turn admit fair classification outcomes.

\section{Adversarially Fair Representations} \label{model}
\newcommand{\arrowWidth}{1.7pt}
\newcommand{\rectWidth}{0.5cm}
\newcommand{\circleSize}{0.9cm}
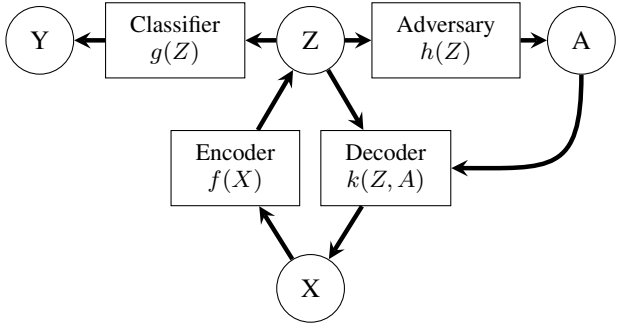
\begin{figure}[ht]
\vskip 0.2in
\begin{center}
\begin{tikzpicture}
\path  (6.6,3.3) node[circle,draw,minimum size=\circleSize,align=center](a) {A}
(3,3.3) node[circle,draw,minimum size=\circleSize,align=center](z) {Z}
(-0.6,3.3) node[circle,draw,minimum size=\circleSize](y) {Y}
(3,0) node[circle,draw,minimum size=\circleSize,align=center](x) {X}
(2,1.6) node[rectangle,draw,minimum width=\rectWidth,minimum height=1cm](encoder) {\footnotesize \begin{tabular}{c} Encoder \\ $f(X)$ \end{tabular}}
(4,1.6) node[rectangle,draw,minimum width=\rectWidth,minimum height=1cm](decoder) {\footnotesize \begin{tabular}{c} Decoder \\ $k(Z, A)$ \end{tabular}}
(1.2,3.3) node[rectangle,draw,minimum width=\rectWidth,minimum height=1cm](classifier) {\footnotesize \begin{tabular}{c} Classifier \\ $g(Z)$ \end{tabular}}
(4.8,3.3) node[rectangle,draw,minimum width=\rectWidth,minimum height=1cm](adversary) {\footnotesize \begin{tabular}{c} Adversary \\ $h(Z)$ \end{tabular}};
\draw[->,>=stealth, line width=\arrowWidth] (x) -- (encoder);
\draw[->,>=stealth, line width=\arrowWidth] (decoder) -- (x);
\draw[->,>=stealth, line width=\arrowWidth] (z) -- (decoder);
\draw[->,>=stealth, line width=\arrowWidth] (encoder) -- (z);
\draw[->,>=stealth, line width=\arrowWidth] (z) -- (classifier);
\draw[->,>=stealth, line width=\arrowWidth] (classifier) -- (y);
\draw[->,>=stealth, line width=\arrowWidth] (z) -- (adversary);
\draw[->,>=stealth, line width=\arrowWidth] (adversary) -- (a);
\draw[->,>=stealth, line width=\arrowWidth] (a) to [out=-90,in=0,looseness=1.6] (decoder);
\end{tikzpicture}
\caption{
    Model for learning adversarially fair representations. The variables are data $X$, latent representations $Z$, sensitive attributes $A$, and labels $Y$. The encoder $f$ maps $X$ (and possibly $A$ - not shown) to $Z$, the decoder $k$ reconstructs $X$ from $(Z, A)$, the classifier $g$ predicts $Y$ from $Z$, and the adversary $h$ predicts $A$ from $Z$ (and possibly $Y$ - not shown). 
}
\label{modelpic}
\end{center}
\vskip -0.2in
\end{figure}

\subsection{A Generalized Model} \label{gen-model}
\iffalse
\begin{samepage}
We assume a generalized model (Figure \ref{modelpic}), which seeks to learn a data representation $Z$ capable of
\begin{itemize}
\item reconstructing the inputs $X$;
\item classifying the target labels $Y$; and
\item protecting the sensitive attribute $A$ from an adversary.
\end{itemize}
\end{samepage}
\else
We assume a generalized model (Figure \ref{modelpic}), which seeks to learn a data representation $Z$ capable of reconstructing the inputs $X$, classifying the target labels $Y$, and protecting the sensitive attribute $A$ from an adversary.
\fi
Either of the first two requirements can be omitted by setting hyperparameters to zero, so the model easily ports to strictly supervised or unsupervised settings as needed.
This general formulation was originally proposed by \citet{edwards2015censoring}; below we address our specific choices of adversarial objectives and explore their fairness implications, which distinguish our work as more closely aligned to the goals of fair representation learning.

The dataset consists of tuples $(X, A, Y)$ in $\mathbb{R}^{n}$, $\{0, 1\}$ and $\{0, 1\}$, respectively.
The encoder $f: \mathbb{R}^{n} \rightarrow \mathbb{R}^m$ yields the representations $Z$.
The encoder can also optionally receive $A$ as input.
The classifier and adversary\footnote{
In learning equalized odds or equal opportunity representations, the adversary $h: \mathbb{R}^{m} \times \{0, 1\} \rightarrow \{0, 1\}$ also takes the label $Y$ as input. 
} $g, h: \mathbb{R}^m \rightarrow \{0, 1\}$ each act on $Z$ and attempt to predict $Y$ and $A$, respectively.
Optionally, a decoder $k: \mathbb{R}^{m} \times \{0, 1\} \rightarrow \mathbb{R}^n$ attempts to reconstruct the original data from the representation and the sensitive variable.

The adversary $h$ seeks to maximize its objective $L_{Adv} (h(f(X, A)), A)$.
We discuss a novel and theoretically motivated adversarial objective in Sections \ref{sec:learning} and \ref{theory}, whose exact terms are modified according to the fairness desideratum.

Meanwhile, the encoder, decoder, and classifier jointly seek to minimize classification loss and reconstruction error, and also minimize the adversary's objective. 
Let $L_C$ denote a suitable classification loss (e.g., cross entropy, $\ell_1$), and $L_{Dec}$ denote a suitable reconstruction loss (e.g., $\ell_2$).
Then we train the generalized model according to the following min-max procedure:
\begin{equation} \label{minmax}
\begin{aligned}
\underset{f, g, k}{\minimize} \medspace &\underset{h}{\maximize} \medspace \E_{X,Y,A} \left[ L(f, g, h, k) \right]
,
\end{aligned}
\end{equation}
with the combined objective expressed as
\begin{equation} \label{eq:objective}
\begin{aligned}
L(f, g, h, k) &= \alpha L_C (g(f(X, A)), Y)\\
&\quad + \beta L_{Dec}(k(f(X, A), A), X)\\
    &\quad\quad+ \gamma L_{Adv} (h(f(X, A)), A)
\end{aligned}
\end{equation}

The hyperparameters $\alpha, \beta, \gamma$ respectively specify a desired balance between utility, reconstruction of the inputs, and fairness.

Due to the novel focus on fair transfer learning, we call our model Learned Adversarially Fair and Transferable Representations (LAFTR).

\subsection{Learning}\label{sec:learning}
We realize $f$, $g$, $h$, and $k$ as neural networks and alternate gradient decent and ascent steps to optimize their parameters according to (\ref{eq:objective}). 
First the encoder-classifier-decoder group ($f,g,k$) takes a gradient step to minimize $L$ while the adversary $h$ is fixed, then $h$ takes a step to maximize $L$ with fixed ($f,g,k$).
Computing gradients necessitates relaxing the binary functions $g$ and $h$, the details of which are discussed in Section \ref{sec:theory-discussion}.

One of our key contributions is a suitable adversarial objective, which we express here and discuss further in Section \ref{theory}.
For shorthand we denote the adversarial objective $L_{Adv} (h(f(X, A)), A)$---whose functional form depends on the desired fairness criteria---as $L_{Adv}(h)$.
For demographic parity, we take the average absolute difference on each sensitive group $\mathcal{D}_0, \mathcal{D}_1$:
\begin{equation}\label{eq:adv-obj}
    L_{Adv}^{DP}(h) = 1 - \sum_{i \in \{0, 1\}} \frac{1}{|\mathcal{D}_i|}\sum_{(x, a) \in \mathcal{D}_i} | h(f(x, a)) - a |\\ 
\end{equation}
For equalized odds, we take the average absolute difference on each sensitive group-label combination $\mathcal{D}_0^0, \mathcal{D}_1^0, \mathcal{D}_0^1, \mathcal{D}_1^1$, where $\mathcal{D}_i^j = \{(x, y, a) \in \mathcal{D} | a = i, y = j\}$:
\begin{equation}\label{eq:adv-eo}
    L_{Adv}^{EO}(h) = 2 - \sum_{(i, j) \in \{0, 1\}^2} \frac{1}{|\mathcal{D}_i^j|}\sum_{(x, a) \in \mathcal{D}_i^j}  | h(f(x, a)) - a |\\ 
\end{equation}
To achieve equal opportunity, we need only sum terms corresponding to $Y = 0$. 

\subsection{Motivation} \label{intuition}
For intuition on this approach and the upcoming theoretical section, we return to the framework from Section \ref{intro}, with a data \textit{owner} who sells representations to a (prediction) \textit{vendor}.
Suppose the data owner is concerned about the unfairness in the predictions made by vendors who use their data. Given that vendors are strategic actors with goals, the owner may wish to guard against two types of vendors:
\begin{itemize}
\item The \textit{indifferent} vendor: this vendor is concerned with utility maximization, and doesn't care about the fairness or unfairness of their predictions.
\item The \textit{adversarial} vendor: this vendor will attempt to actively discriminate by the sensitive attribute.
\end{itemize}
In the adversarial model defined in Section \ref{gen-model}, the encoder is what the data owner really wants; this yields the representations which will be sold to vendors.
When the encoder is learned, the other two parts of the model ensure that the representations respond appropriately to each type of vendor: the classifier ensures utility by simulating an indifferent vendor with a prediction task, and the adversary ensures fairness by simulating an adversarial vendor with discriminatory goals.
It is important to the data owner that the model's adversary be as strong as possible---if it is too weak, the owner will underestimate the unfairness enacted by the adversarial vendor.

However, there is another important reason why the model should have a strong adversary, which is key to our theoretical results.
Intuitively, the degree of unfairness achieved by the adversarial vendor (who is optimizing for unfairness) will not be exceeded by the indifferent vendor.
Beating a strong adversary $h$ during training implies that downstream classifiers naively trained on the learned representation $Z$ must also act fairly.
Crucially, this fairness bound depends on the discriminative power of $h$; this motivates our use of the representation $Z$ as a direct input to $h$, because it yields a strictly more powerful $h$ and thus tighter unfairness bound than adversarially training on the predictions and labels alone as in \citet{zhang2018mitigating}.

\section{Theoretical Properties} \label{theory}
We now draw a connection between our choice of adversarial objective and several common metrics from the fair classification literature.
We derive adversarial upper bounds on unfairness that can be used in adversarial training to achieve either demographic parity, equalized odds, or equal opportunity.

We are interested in quantitatively comparing two distributions corresponding to the learned group representations, so consider two distributions $\mathcal{D}_0$ and $\mathcal{D}_1$ over the same sample space $\Omega_{\mathcal{D}}$, as well as a binary test function $\mu: \Omega_{\mathcal{D}} \rightarrow \{0, 1\}$. 
$\mu$ is called a test since it can distinguish between samples from the two distributions according to the absolute difference in its expected value.
We call this quantity the \emph{test discrepancy} and express it as
\begin{equation}
    d_\mu(\mathcal{D}_0, \mathcal{D}_1) \triangleq | \underset{x \sim \mathcal{D}_0}{\E} \left[ \mu(x) \right] - \underset{x \sim \mathcal{D}_1}{\E} \left[ \mu(x) \right] |
.
\end{equation}
The \textit{statistical distance} (a.k.a. total variation distance) between distributions is defined as the maximum attainable test discrepancy \citep{cover2012elements}:
\begin{equation} \label{stat_dist}
\begin{aligned}
    \Delta^*(\mathcal{D}_0, \mathcal{D}_1) \triangleq \sup_{\mu} d_\mu(\mathcal{D}_0, \mathcal{D}_1)
.
\end{aligned}
\end{equation}

When learning fair representations we are interested in the distribution of $Z$ conditioned on a specific value of group membership $A \in \{0, 1\}$.
As a shorthand we denote the distributions $p(Z|A=0)$ and $p(Z|A=1)$ as $\mathcal{Z}_0$ and $\mathcal{Z}_1$, respectively.

\subsection{Bounding Demographic Parity}\label{theory-dem-par}
In supervised learning we seek a $g$ that accurately predicts some label $Y$; in fair supervised learning we also want to quantify $g$ according to the fairness metrics discussed in Section \ref{sec:background}.
For example, the demographic parity distance is expressed as the absolute expected difference in classifier outcomes between the two groups:
\begin{equation}
    \Delta_{DP}(g) \triangleq d_{g}(\mathcal{Z}_0,\mathcal{Z}_1) = | \E_{\mathcal{Z}_0}[g] - \E_{\mathcal{Z}_1}[g] |
    .
\end{equation}
Note that $\Delta_{DP}(g) \leq \Delta^*(\mathcal{Z}_0,\mathcal{Z}_1)$, and also that $\Delta_{DP}(g) = 0$ if and only if $g(Z) \independent A$, i.e., demographic parity has been achieved.

Now consider an adversary $h: \Omega_{\mathcal{Z}} \rightarrow \{0, 1\}$ whose objective (negative loss) function\footnote{
    This is equivalent to the objective expressed by Equation \ref{eq:adv-obj} when the expectations are evaluated with finite samples.    
} is expressed as 
\begin{equation}\label{eq:adv-reward}
    L_{Adv}^{DP}(h) \triangleq \E_{\mathcal{Z}_0}[1-h] + \E_{\mathcal{Z}_1}[h] - 1 
    .
\end{equation}
Given samples from $\Omega_{\mathcal{Z}}$ the adversary seeks to correctly predict the value of $A$, and learns by maximizing $L_{Adv}^{DP}(h)$.
Given an optimal adversary trained to maximize (\ref{eq:adv-reward}), the adversary's loss will bound $\Delta_{DP}(g)$ from above for any function $g$ learnable from $Z$.
Thus a sufficiently powerful adversary $h$ will expose through the value of its objective the demographic disparity of any classifier $g$.
We will later use this to motivate learning fair representations via an encoder $f: \mathcal{X} \rightarrow \mathcal{Z}$ that simultaneously minimizes the task loss and minimizes the adversary objective.

\textbf{Theorem.} \textit{
Consider a classifier $g:\Omega_{\mathcal{Z}}\rightarrow \Omega_{\mathcal{Y}}$ and adversary $h:\Omega_{\mathcal{Z}}\rightarrow \Omega_{\mathcal{A}}$ as binary functions, i.e,. $\Omega_{\mathcal{Y}}=\Omega_{\mathcal{A}}= \{0, 1\}$. 
Then $L_{Adv}^{DP}(h^*) \geq \Delta_{DP}(g)$: the demographic parity distance of $g$ is bounded above by the optimal objective value of $h$.
}

\textbf{Proof.} 
By definition $\Delta_{DP}(g) \geq 0$.
Suppose without loss of generality (WLOG) that $\E_{\mathcal{Z}_0}[g] \geq \E_{\mathcal{Z}_1}[g]$, i.e., the classifier predicts the ``positive'' outcome more often for group $A_0$ in expectation. 
Then, an immediate corollary is $\E_{\mathcal{Z}_1}[1-g] \geq \E_{\mathcal{Z}_0}[1-g]$, and we can drop the absolute value in our expression of the disparate impact distance:
\begin{equation}
\begin{aligned}
\Delta_{DP}(g) = \E_{\mathcal{Z}_0}[g] - \E_{\mathcal{Z}_1}[g] = \E_{\mathcal{Z}_0}[g] + \E_{\mathcal{Z}_1}[1 - g] - 1 
\end{aligned}
\end{equation}
where the second equality is due to $\E_{\mathcal{Z}_1}[g] = 1 - \E_{\mathcal{Z}_1}[1-g]$.
Now consider an adversary that guesses the opposite of $g$
, i.e., $h = 1 - g$. 
Then\footnote{
Before we assumed WLOG $\E_{\mathcal{Z}_0}[g] \geq \E_{\mathcal{Z}_1}[g]$. 
If instead $\E_{\mathcal{Z}_0}[g] < \E_{\mathcal{Z}_1}[g]$ then we simply choose $h = g$ instead to achieve the same result.
}, we have 
\begin{equation} \label{tight-DP-bound}
\begin{aligned}
    L_{Adv}^{DP}(h) = L_{Adv}^{DP}(1-g) &= \E_{\mathcal{Z}_0}[g] + \E_{\mathcal{Z}_1}[1-g] - 1 \\
    &= \Delta_{DP}(g)
\end{aligned}
\end{equation}
The optimal adversary $h^\star$ does at least as well as any arbitrary choice of $h$, therefore $L_{Adv}^{DP}(h^\star) \geq L_{Adv}^{DP}(h) = \Delta_{DP}$.
\hfill$\blacksquare$

\subsection{Bounding Equalized Odds}\label{theory-eq-odds}
We now turn our attention to equalized odds. 
First we extend our shorthand to denote $p(Z| A=a, Y=y)$ as $\mathcal{Z}_{a}^{y}$, the representation of group $a$ conditioned on a specific label $y$.
The equalized odds distance of classifier $g: \Omega_{\mathcal{Z}} \rightarrow \{0, 1\}$ is
\begin{equation}\label{eq:delta-eo}
    \begin{aligned}
        \Delta_{EO}(g) &\triangleq | \E_{\mathcal{Z}_0^0}[g] - \E_{\mathcal{Z}_1^0}[g] | \\
        &+ | \E_{\mathcal{Z}_0^1}[1 - g] - \E_{\mathcal{Z}_1^1}[1 - g] |
,
    \end{aligned}
\end{equation}
which comprises the absolute difference in false positive rates plus the absolute difference in false negative rates.
$\Delta_{EO}(g) = 0$ means $g$ satisfies equalized odds. 
Note that $\Delta_{EO}(g) \leq \Delta(\mathcal{Z}_0^0, \mathcal{Z}_1^0) + \Delta(\mathcal{Z}_0^1, \mathcal{Z}_1^1)$. 

We can make a similar claim as above for equalized odds: given an optimal adversary trained on $Z$ with the appropriate objective, if the adversary also receives the label $Y$, the adversary's loss will upper bound $\Delta_{EO}(g)$ for any function $g$ learnable from $Z$.

\textbf{Theorem.} \textit{
Let the classifier $g: \Omega_{\mathcal{Z}} \rightarrow \Omega_{\mathcal{Y}}$ and the adversary $h: \Omega_{\mathcal{Z}} \times \Omega_{\mathcal{Y}} \rightarrow \Omega_{\mathcal{Z}}$, as binary functions, i.e., $\Omega_{\mathcal{Y}} = \Omega_{\mathcal{A}} = \{0, 1\}$.
Then $L_{Adv}^{EO}(h^*) \geq \Delta_{EO}(g)$: the equalized odds distance of $g$ is bounded above by the optimal objective value of $h$.
}    

\textbf{Proof.} Let the adversary $h$'s objective be
\begin{equation}
\begin{aligned}
    L_{Adv}^{EO}(h) &= \E_{\mathcal{Z}_0^0}[1 - h] + \E_{\mathcal{Z}_1^0}[h]\\
&+ \E_{\mathcal{Z}_0^1}[1 - h] + \E_{\mathcal{Z}_1^1}[h] - 2\\
\end{aligned}
\end{equation}

By definition $\Delta_{EO}(g) \geq 0$.
Let  $| \E_{\mathcal{Z}_0^0}[g] - \E_{\mathcal{Z}_1^0}[g] | = \alpha \in [0, \Delta_{EO}(g)]$ and $| \E_{\mathcal{Z}_0^1}(1 - g) - \E_{\mathcal{Z}_1^1}[1 - g] | = \Delta_{EO}(g) - \alpha$. 
WLOG, suppose $\E_{\mathcal{Z}_0^0}[g] \geq \E_{\mathcal{Z}_1^0}[g]$ and $\E_{\mathcal{Z}_0^1}[1 - g] \geq \E_{\mathcal{Z}_1^1}[1 - g]$. 
Thus we can partition (\ref{eq:delta-eo}) as two expressions, which we write as 
\begin{equation}
\begin{aligned}
\E_{\mathcal{Z}_0^0}[g] + \E_{\mathcal{Z}_1^0}[1 - g] &= 1 + \alpha,\\
\E_{\mathcal{Z}_0^1}[1 - g] + \E_{\mathcal{Z}_1^1}[g] &= 1 + (\Delta_{EO}(g) - \alpha),\\
\end{aligned}
\end{equation}
which can be derived using the familiar identity $\E_{p}[\eta] = 1-\E_{p}[1-\eta]$ for binary functions.

Now, let us consider the following adversary $h$
\begin{equation}
\begin{aligned}
    h(z) = \left\{\begin{array}{lr}
        g(z), & \text{if } y = 1\\
        1 - g(z), & \text{if } y = 0
        \end{array}\right\} 
        .
\end{aligned}
\end{equation}
Then the previous statements become
\begin{equation}
\begin{aligned}
\E_{\mathcal{Z}_0^0}[1 - h] + \E_{\mathcal{Z}_1^0}[h] &= 1 + \alpha\\
\E_{\mathcal{Z}_0^1}[1 - h] + \E_{\mathcal{Z}_1^1}[h] &= 1 + (\Delta_{EO}(g) - \alpha)\\
\end{aligned}
\end{equation}
Recalling our definition of $L_{Adv}^{EO}(h)$, this means that 
\begin{equation}
\begin{aligned}
    L_{Adv}^{EO}(h) &= \E_{\mathcal{Z}_0^0}[1 - h] + \E_{\mathcal{Z}_1^0}[h] + \E_{\mathcal{Z}_0^1}[h] + \E_{\mathcal{Z}_1^1}[h] - 2 \\
&= 1 + \alpha + 1 + (\Delta_{EO}(g) - \alpha) - 2 = \Delta_{EO}(g)\\
\end{aligned}
\end{equation}
That means that for the optimal adversary $h^\star = \sup_h L_{Adv}^{EO}(h)$, we have $L_{Adv}^{EO}(h^\star) \geq L_{Adv}^{EO}(h) = \Delta_{EO}$.
\hfill$\blacksquare$

An adversarial bound for equal opportunity distance, defined as $\Delta_{EOpp}(g) \triangleq | \E_{\mathcal{Z}_0^0}[g] - \E_{\mathcal{Z}_1^0}[g] |$, can be derived similarly.

\subsection{Additional points}\label{sec:theory-discussion}
One interesting note is that in each proof, we provided an example of an adversary which was calculated only from the joint distribution of $Y, A$, and $\hat Y = g(Z)$---we did not require direct access to $Z$---and this adversary achieved a loss exactly equal to the quantity in question ($\Delta_{DP}$ or $\Delta_{EO}$). Therefore, if we only allow our adversary access to those outputs, our adversarial objective (assuming an optimal adversary), is equivalent to simply adding either $\Delta_{DP}$ or $\Delta_{EO}$ to our classification objective, similar to common regularization approaches  \citep{kamishima2012fairness,bechavod2017learning,madras2017predict,zafar2017fairness}.
Below we consider a stronger adversary, with direct access to the key intermediate learned representation $Z$. This allows for a potentially greater upper bound for the degree of unfairness, which in turn forces any classifier trained on $Z$ to act fairly.

In our proofs we have considered the classifier $g$ and adversary $h$ as binary functions.
In practice we want to learn these functions by gradient-based optimization, so we instead substitute their continuous relaxations $\tilde g, \tilde h: \Omega_{\mathcal{Z}} \rightarrow [0,1]$.
By viewing the continuous output as parameterizing a Bernoulli distribution over outcomes we can follow the same steps in our earlier proofs to show that in both cases (demographic parity and equalized odds)  $\E[L(\bar h^*)] \geq \E[\Delta(\bar g)]$, where $\bar h^*$ and $\bar g$ are randomized binary classifiers parameterized by the outputs of $\tilde h^*$ and $\tilde g$.

\subsection{Comparison to \citet{edwards2015censoring}} \label{sec:compare-storkey}
An alternative to optimizing the expectation of the randomized classifier $\tilde h$ is to minimize its negative log likelihood (NLL - also known as cross entropy loss), given by
\begin{equation}\label{eq:bernoulli}
L(\tilde h) = -\E_{Z,A} \left[ A \log \tilde h(Z) + (1-A) \log (1-\tilde h(Z)) \right]
.
\end{equation}
This is the formulation adopted by
\citet{ganin2016domain} and \citet{edwards2015censoring}, which propose maximizing (\ref{eq:bernoulli}) as a proxy for computing the statistical distance\footnote{
These papers discuss the bound on $\Delta_{DP}(g)$ in terms of the $\mathcal{H}$-divergence \cite{blitzer2006domain}, which is simply the statistical distance $\Delta^*$ up to a multiplicative constant.
}
$\Delta^*(\mathcal{Z}_0, \mathcal{Z}_1)$ during adversarial training. 

The adversarial loss we adopt here instead of cross-entropy is group-normalized $\ell_1$, defined in Equations $\ref{eq:adv-obj}$ and $\ref{eq:adv-eo}$.
The main problems with cross entropy loss in this setting arise from the fact that the adversarial objective should be calculating the test discrepancy.
However, the cross entropy objective sometimes fails to do so, for example when the dataset is imbalanced.
In Appendix \ref{sec:app-ce}, we discuss a synthetic example where a cross-entropy loss will incorrectly guide an adversary on an unbalanced dataset, but a group-normalized $\ell_1$ adversary will work correctly.

Furthermore, group normalized $\ell_1$ corresponds to a more natural relaxation of the fairness metrics in question.
It is important that the adversarial objective incentivizes the test discrepancy, as group-normalized $\ell_1$ does; this encourages the adversary to get an objective value as close to $\Delta^\star$ as possible, which is key for fairness (see Section \ref{intuition}).
In practice, optimizing $\ell_1$ loss with gradients can be difficult, so while we suggest it as a suitable theoretically-motivated continuous relaxation for our model (and present experimental results), there may be other suitable options beyond those considered in this work.

\section{Experiments} \label{experiments}

\subsection{Fair classification}\label{sec:fair-classification}
\begin{figure*}[ht!]
\centering
\begin{subfigure}[t]{0.33\textwidth}
\centering
\includegraphics[width=\textwidth]{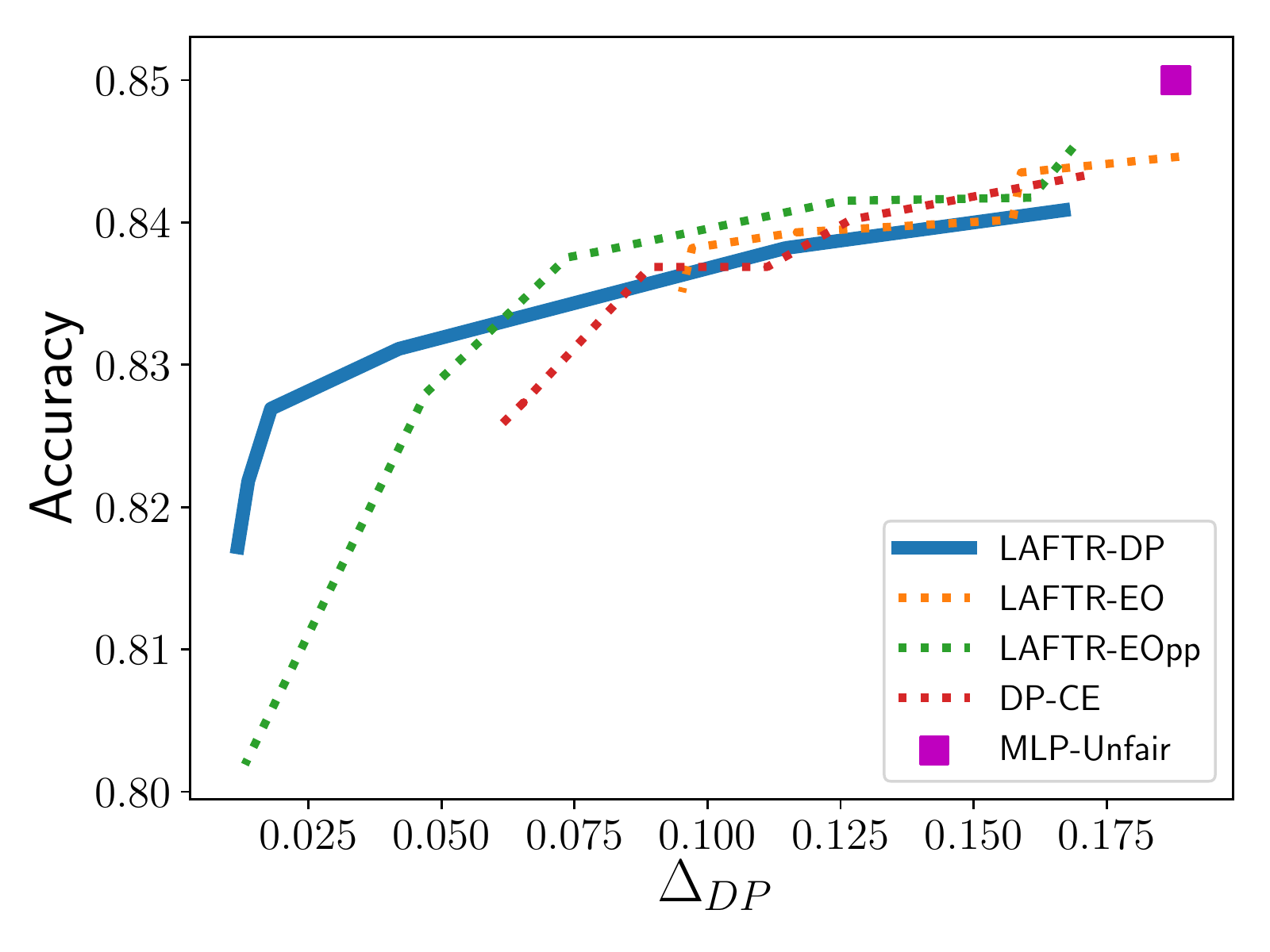}
\caption{Tradeoff between accuracy and $\Delta_{DP}$}
\label{results:pareto-DP}
\end{subfigure}%
\hfill
\begin{subfigure}[t]{0.33\textwidth}
\centering
\includegraphics[width=\textwidth]{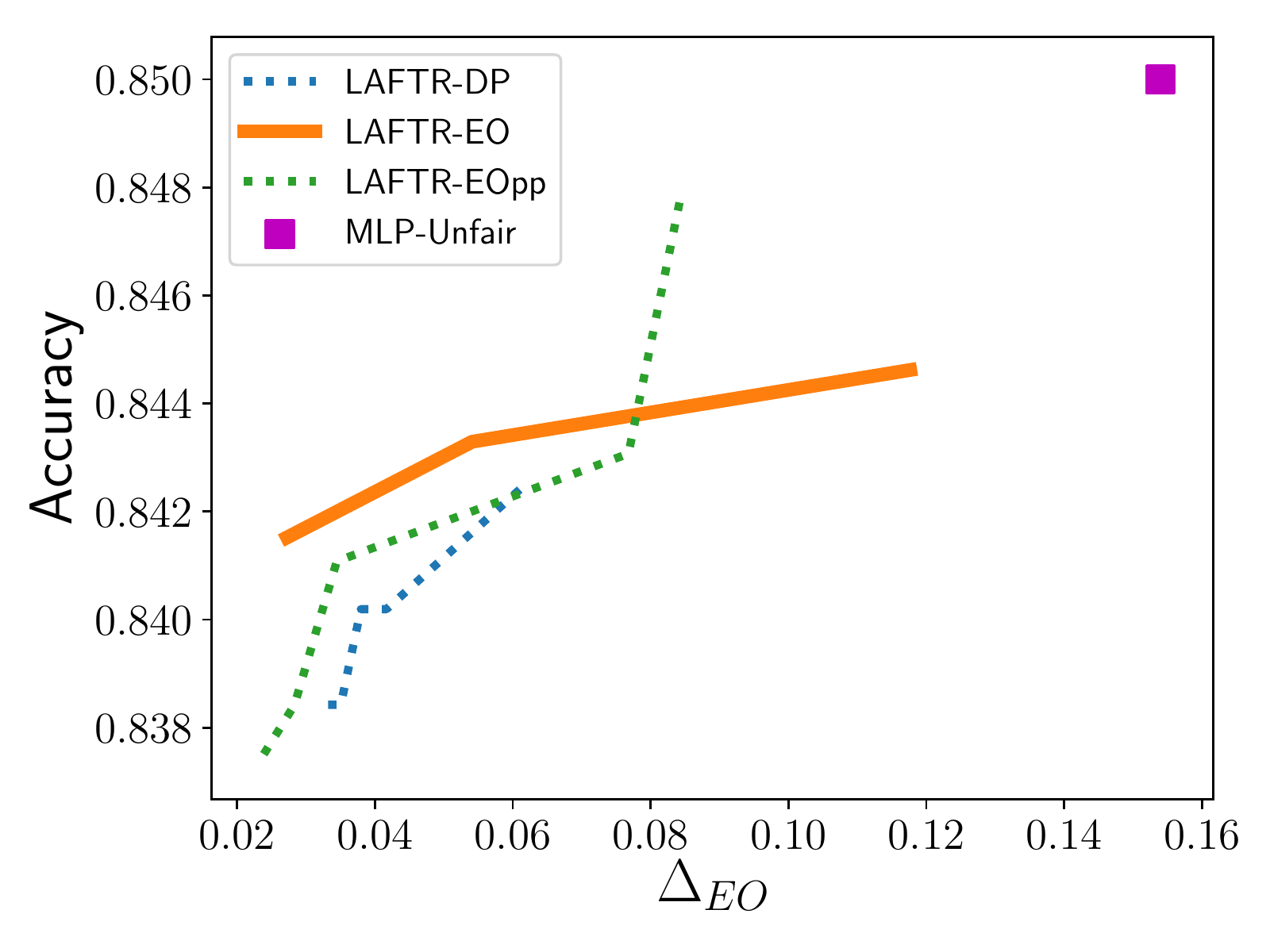}
\caption{Tradeoff between accuracy and $\Delta_{EO}$}
\label{results:pareto-EO}
\end{subfigure}%
\hfill
\begin{subfigure}[t]{0.33\textwidth}
\centering
\includegraphics[width=\textwidth]{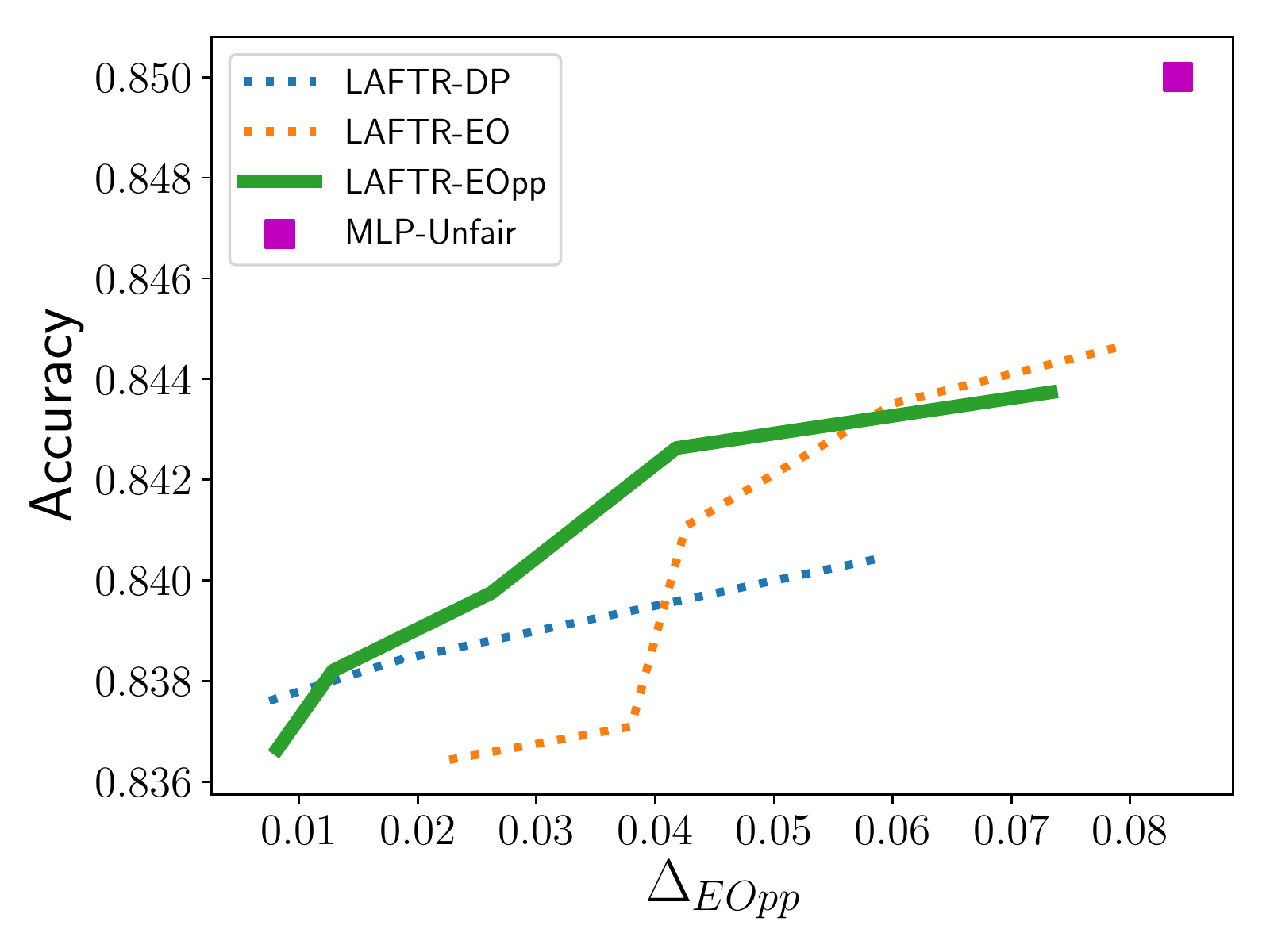}
\caption{Tradeoff between accuracy and $\Delta_{EOpp}$}
\label{results:pareto-EOpp}
\end{subfigure}%
    \caption{
        Accuracy-fairness tradeoffs for various fairness metrics ($\Delta_{DP}$, $\Delta_{EO}$, $\Delta_{EOpp}$), and LAFTR adversarial objectives $(L_{Adv}^{DP}, L_{Adv}^{EO}, L_{Adv}^{EOpp})$ on fair classification of the Adult dataset.
Upper-left corner (high accuracy, low $\Delta$) is preferable.
Figure \ref{results:pareto-DP} also compares to a cross-entropy adversarial objective \citep{edwards2015censoring}, denoted DP-CE.
Curves are generated by sweeping a range of fairness coefficients $\gamma$, taking the median across 7 runs per $\gamma$, and computing the Pareto front.
In each plot, the bolded line is the one we expect to perform the best.
Magenta square is a baseline MLP with no fairness constraints.
see Algorithm \ref{alg:laftr} and Appendix \ref{sec:training-details}.
    }
\label{results:pareto-fairness}

\end{figure*}

LAFTR seeks to learn an encoder yielding fair representations, i.e., the encoder's outputs can be used by third parties with the assurance that their naively trained classifiers will be reasonably fair and accurate.
Thus we evaluate the quality of the encoder according to the following training procedure, also described in pseudo-code by Algorithm 1.
Using labels $Y$, sensitive attribute $A$, and data $X$, we train an encoder using the adversarial method outlined in Section \ref{model}, receiving both $X$ and $A$ as inputs.
We then freeze the learned encoder; from now on we use it only to output representations $Z$. 
Then, using unseen data, we train a classifier on top of the frozen encoder. The classifier learns to predict $Y$ from $Z$ --- note, this classifier is not trained to be fair. 
We can then evaluate the accuracy and fairness of this classifier on a test set to assess the quality of the learned representations.

During Step 1 of Algorithm 1, the learning algorithm is specified either as a baseline (e.g., unfair MLP) or as LAFTR, i.e., stochastic gradient-based optimization of (Equation \ref{minmax}) with one of the three adversarial objectives described in Section \ref{sec:learning}.
When LAFTR is used in Step 1, all but the encoder $f$ are discarded in Step 2.
For all experiments we use cross entropy loss for the classifier (we observed training unstability with other classifier losses).
The classifier $g$ in Step 3 is a feed-forward MLP trained with SGD.
See Appendix \ref{sec:training-details} for details.
\begin{algorithm}[tb]\captionsetup{labelfont={sc,bf}}
    \caption{Evaluation scheme for fair classification ($Y'=Y$) \& transfer learning ($Y' \neq Y$).}
   \label{alg:laftr}
\begin{algorithmic}
   \STATE {\bfseries Input:} data $X \in \Omega_X$, sensitive attribute $A \in \Omega_A$, labels $Y, Y' \in \Omega_Y$, representation space $\Omega_Z$ 
   \STATE {\bfseries Step 1:} Learn an encoder $f: \Omega_X \rightarrow \Omega_Z$ using data $X$, task label $Y$, and sensitive attribute $A$.
   \STATE {\bfseries Step 2:} Freeze $f$.
   \STATE {\bfseries Step 3:} Learn a classifier (without fairness constraints) $g: \Omega_Z \rightarrow \Omega_Y$ on top of $f$, using  data $f(X')$, task label $Y'$, and sensitive attribute $A'$.
   \STATE {\bfseries Step 3:} Evaluate the fairness and accuracy of the composed classifier $g \circ f: \Omega_X \rightarrow \Omega_Y$ on held out test data, for task $Y'$.
\end{algorithmic}
\end{algorithm}

We evaluate the performance of our model\footnote{See \href{https://github.com/VectorInstitute/laftr}{https://github.com/VectorInstitute/laftr} for code.
}on fair classification on the UCI Adult dataset\footnote{https://archive.ics.uci.edu/ml/datasets/adult}, which contains over 40,000 rows of information describing adults from the 1994 US Census. 
We aimed to predict each person's income category (either greater or less than 50K/year). 
We took the sensitive attribute to be gender, which was listed as Male or Female. 

Figure \ref{results:pareto-fairness} shows classification results on the Adult dataset.
Each sub-figure shows the accuracy-fairness trade-off (for varying values of $\gamma$; we set $\alpha = 1, \beta = 0$ for all classification experiments) evaluated according to one of the group fairness metrics: $\Delta_{DP}$, $\Delta_{EO}$, and $\Delta_{EOpp}$.
For each fairness metric, we show the trade-off curves for LAFTR trained under three adversarial objectives: $L_{Adv}^{DP}$, $L_{Adv}^{EO}$, and $L_{Adv}^{EOpp}$.
We observe, especially in the most important regiment for fairness (small $\Delta$), that the adversarial objective we propose for a particular fairness metric tends to achieve the best trade-off.
Furthermore, in Figure \ref{results:pareto-DP}, we compare our proposed adversarial objective for demographic parity with the one proposed in \citep{edwards2015censoring}, finding a similar result.

For low values of un-fairness, i.e., minimal violations of the respective fairness criteria, the LAFTR model trained to optimize the target criteria obtains the highest test accuracy.
While the improvements are somewhat uneven for other regions of fairness-accuracy space (which we attribute to instability of adversarial training), this demonstrates the potential of our proposed objectives.
However, the fairness of our model's learned representations are not limited to the task it is trained on.
We now turn to experiments which demonstrate the utility of our model in learning fair representations for a variety of tasks.

\subsection{Transfer Learning} \label{sec:transfer}
In this section, we show the promise of our model for \textit{fair transfer learning}.
As far as we know, beyond a brief introduction in \citet{zemel2013learning}, we provide the first in-depth experimental results on this task, which is pertinent to the common situation where the data owner and vendor are separate entities.

We examine the Heritage Health dataset\footnote{https://www.kaggle.com/c/hhp}, which comprises insurance claims and physician records relating to the health and hospitalization of over 60,000 patients.
We predict the Charlson Index, a comorbidity indicator that estimates the risk of patient death in the next several years. 
We binarize the (nonnegative) Charlson Index as zero/nonzero. 
We took the sensitive variable as binarized age (thresholded at 70 years old). 
This dataset contains information on sex, age, lab test, prescription, and claim details. 

The task is as follows: using data $X$, sensitive attribute $A$, and labels $Y$, learn an encoding function $f$ such that given unseen $X'$, the representations produced by $f(X', A)$ can be used to learn a fair predictor for new task labels $Y'$, even if the new predictor is being learned by a vendor who is indifferent or adversarial to fairness.
This is an intuitively desirable condition: if the data owner can guarantee that predictors learned from their representations will be fair, then there is no need to impose fairness restrictions on vendors, or to rely on their goodwill.

The original task is to predict Charlson index $Y$ fairly with respect to age $A$.
The transfer tasks relate to the various primary condition group (PCG) labels, each of which indicates a patient's insurance claim corresponding to a specific medical condition.
PCG labels $\{Y'\}$ were held out during LAFTR training but presumably correlate to varying degrees with the original label $Y$.
The prediction task was binary: did a patient file an insurance claim for a given PCG label in this year? For various patients, this was true for zero, one, or many PCG labels. 
There were 46 different PCG labels in the dataset; we considered only used the 10 most common---whose positive base rates ranged from 9-60\%---as transfer tasks.

Our experimental procedure was as follows. 
To learn representations that transfer fairly, we used the same model as described above, but set our reconstruction coefficient $\beta = 1$.
Without this, the adversary will stamp out any information not relevant to the label from the representation, which will hurt transferability.
We can optionally set our classification coefficient $\alpha$ to 0, which worked better in practice.
Note that although the classifier $g$ is no longer involved when $\alpha = 0$, the target task labels are still relevant for either equalized odds or equal opportunity transfer fairness.

We split our test set ($\sim20,000$ examples) into transfer-train, -validation, and -test sets. 
We trained LAFTR ($\alpha = 0, \beta = 1$, $\ell_2$ loss for the decoder) on the full training set, and then only kept the encoder. 
In the results reported here, we trained using the equalized odds adversarial objective described in Section \ref{sec:learning}; similar results were obtained with the other adversarial objectives.
Then, we created a feed-forward model which consisted of our frozen, adversarially-learned encoder followed by an MLP with one hidden layer, with a loss function of cross entropy with no fairness modifications. 
Then, $\forall \ i \in 1 \dots 10$, we trained this feed-forward model on PCG label $i$ (using the transfer-train and -validation) sets, and tested it on the transfer-test set.
This procedure is described in Algorithm \ref{alg:laftr}, with $Y'$ taking 10 values in turn, and $Y$ remaining constant ($Y \neq Y'$).

We trained four models to test our method against. 
The first was an MLP predicting the PCG label directly from the data (Target-Unfair), with no separate representation learning involved and no fairness criteria in the objective---this provides an effective upper bound for classification accuracy. 
The others all involved learning separate representations on the original task, and freezing the encoder as previously described; the internal representations of MLPs have been shown to contain useful information \cite{hinton2006reducing}.
These (and LAFTR) can be seen as the values of \textsc{ReprLearn} in Alg. \ref{alg:laftr}.
In two models, we learned the original $Y$ using an MLP (one regularized for fairness \citep{bechavod2017learning}, one not; Transfer-Fair and -Unfair, respectively) and trained for the transfer task on its internal representations. 
As a third baseline, we trained an adversarial model similar to the one proposed in \cite{zhang2018mitigating}, where the adversary has access only to the classifier output $\hat{Y} = g(Z)$ and the ground truth label (Transfer-Y-Adv), to investigate the utility of our adversary having access to the underlying representation, rather than just the joint classification statistics $(Y, A, \hat{Y})$.

\begin{figure}[ht!]
\vskip 0.2in
\begin{center}
\centerline{\includegraphics[width=\columnwidth]{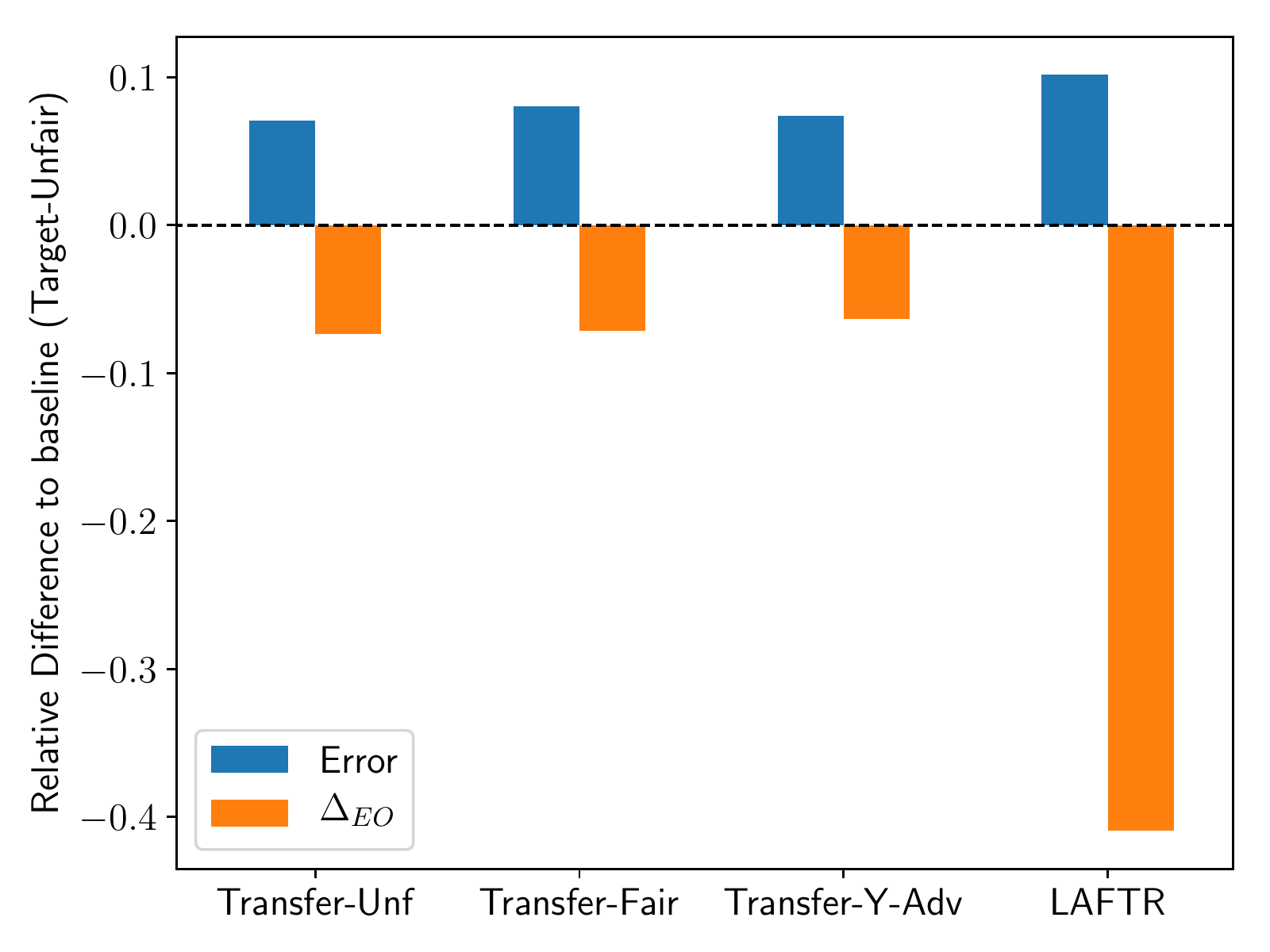}}
\caption{Fair transfer learning on Health dataset. Displaying average across 10 transfer tasks of relative difference in error and $\Delta_{EO}$ unfairness (the lower the better for both metrics), as compared to a baseline unfair model learned directly from the data. -0.10 means a 10\% decrease. Transfer-Unf and -Fair are MLP's with and without fairness restrictions respectively, Transfer-Y-Adv is an adversarial model with access to the classifier output rather than the underlying representations, and LAFTR is our model trained with the adversarial equalized odds objective.}
\label{transfer-learn-results-pic}
\end{center}
\vskip -0.2in
\end{figure}

    We report our results in Figure \ref{transfer-learn-results-pic} and Table \ref{health-transfer}.
In Figure \ref{transfer-learn-results-pic}, we show the relative change from the high-accuracy baseline learned directly from the data for both classification error and $\Delta_{EO}$.
LAFTR shows a clear improvement in fairness; it improves $\Delta_{EO}$ on average from the non-transfer baseline, and the relative difference is an average of $\sim$20\%, which is much larger than other baselines.
We also see that LAFTR's loss in accuracy is only marginally worse than other models.

A fairly-regularized MLP (``Transfer-Fair'') does not actually produce fair representations during trasnfer; on average it yields similar fairness results to transferring representations learned without fairness constraints.
Another observation is that the output-only adversarial model (``Transfer Y-Adv'') produces similar transfer results to the regularized MLP.
This shows the practical gain of using an adversary that can observe the representations.

\begin{table}[!htbp]
\caption{Results from Figure \ref{transfer-learn-results-pic} broken out by task. $\Delta_{EO}$ for each of the 10 transfer tasks is shown, which entails identifying a primary condition code that refers to a particular medical condition. Most fair on each task is bolded. All model names are abbreviated from Figure \ref{transfer-learn-results-pic}; ``TarUnf'' is a baseline, unfair predictor learned directly from the target data without a fairness objective.}
\vskip 0.15in
\begin{center}
\begin{small}
\begin{sc}
\tabcolsep=0.08cm
\begin{tabular}{cccccc}
\toprule
Tra. Task  & TarUnf & TraUnf & TraFair & TraY-AF& LAFTR \\
\midrule
MSC2a3 & 0.362 & 0.370 &  0.381 & 0.378 & \textbf{0.281}\\
METAB3 & 0.510 & 0.579 &  \textbf{0.436} & 0.478 &0.439\\
ARTHSPIN & 0.280 & 0.323 &  0.373 & 0.337 &\textbf{0.188}\\
NEUMENT & 0.419 & 0.419 &  0.332 & 0.450 &\textbf{0.199}\\
RESPR4 & 0.181 & 0.160 &  0.223 & 0.091 &\textbf{0.051}\\
MISCHRT & 0.217 & 0.213 &  0.171 & 0.206 &\textbf{0.095}\\
SKNAUT & 0.324 & \textbf{0.125} &  0.205 & 0.315 & 0.155\\
GIBLEED & 0.189 & 0.176 &  0.141 & 0.187 & \textbf{0.110}\\
INFEC4 & 0.106 & 0.042 &  0.026 & \textbf{0.012} & 0.044\\
TRAUMA & 0.020 & 0.028 & 0.032 & 0.032 & \textbf{0.019}\\
\bottomrule
\end{tabular}
\end{sc}
\end{small}
\end{center}
\label{health-transfer}
\vskip -0.1in
\end{table}

Since transfer fairness varied much more than accuracy, we break out the results of Fig. \ref{transfer-learn-results-pic} in Table \ref{health-transfer}, showing the fairness outcome of each of the 10 separate transfer tasks.
We note that LAFTR provides the fairest predictions on 7 of the 10 tasks, often by a wide margin, and is never too far behind the fairest model for each task.
The unfair model TraUnf achieved the best fairness on one task.
We suspect this is due to some of these tasks being relatively easy to solve without relying on the sensitive attribute by proxy.
Since the equalized odds metric is better aligned with accuracy than demographic parity \citep{hardt2016equality}, high accuracy classifiers can sometimes achieve good $\Delta_{EO}$ if they do not rely on the sensitive attribute by proxy. 
Because the data owner has no knowledge of the downstream task, however, our results suggest that using LAFTR is safer than using the raw inputs;
LAFTR is relatively fair even when TraUnf is the most fair, whereas TraUnf is dramatically less fair than LAFTR on several tasks.

We provide coarser metrics of fairness for our representations in Table \ref{health-transfer-other}. We give two metrics: maximum mean discrepancy (MMD) \citep{gretton2007kernel}, which is a general measure of distributional distance; and adversarial accuracy (if an adversary is given these representations, how well can it learn to predict the sensitive attribute?). 
In both metrics, our representations are more fair than the baselines. We give two versions of the ``Transfer-Y-Adv'' adversarial model ($\beta = 0, 1$); note that it has much better MMD when the reconstruction term is added, but that this does not improve its adversarial accuracy, indicating that our model is doing something more sophisticated than simply matching moments of distributions.

\begin{table}[t]
\caption{Transfer fairness, other metrics. Models are as defined in Figure \ref{transfer-learn-results-pic}.  MMD is calculated with a Gaussian RBF kernel ($\sigma = 1$). AdvAcc is the accuracy of a separate MLP trained on the representations to predict the sensitive attribute; due to data imbalance an adversary predicting 0 on each case obtains accuracy of approximately 0.74.}
\vskip 0.15in
\begin{center}
\begin{small}
\begin{sc}
\begin{tabular}{ccc}
\toprule
Model & MMD & AdvAcc \\
\midrule
Transfer-Unfair & $1.1 \times 10^{-2}$  & 0.787\\
Transfer-Fair & $1.4 \times 10^{-3}$ & 0.784 \\
Transfer-Y-Adv ($\beta = 1$) & $3.4 \times 10^{-5}$ & 0.787 \\
Transfer-Y-Adv ($\beta = 0$) & $1.1 \times 10^{-3}$ & 0.786 \\
LAFTR & $\mathbf{2.7 \times 10^{-5}}$ & \textbf{0.761}\\
\bottomrule
\end{tabular}
\end{sc}
\end{small}
\end{center}
\label{health-transfer-other}
\vskip -0.1in
\end{table}

\section{Conclusion} \label{conclusion}
In this paper, we proposed and explore methods of learning adversarially fair representations. 
We provided theoretical grounding for the concept, and proposed novel adversarial objectives that guarantee performance on commonly used metrics of group fairness. 
Experimentally, we demonstrated that these methods can learn fair and useful predictors through using an adversary on the intermediate representation. 
We also demonstrate success on fair transfer learning, by showing that our methods can produce representations which transfer utility to new tasks as well as yielding fairness improvements.

Several open problems remain around the question of learning representations fairly. 
Various approaches have been proposed, both adversarial and non-adversarial (such as MMD). 
A careful in-depth comparison of these approaches would help elucidate their pros and cons.
Furthermore, questions remain about the optimal form of adversarial loss function, both in theory and practice.
Answering these questions could help stabilize adversarial training of fair representations.
As for transfer fairness, it would be useful to understand between what tasks and in what situations transfer fairness is possible, and on what tasks it is most likely to succeed.

\section*{Acknowledgements}
We gratefully acknowledge Cynthia Dwork and Kevin Swersky for their helpful comments. This work was supported by the Canadian Institute for Advanced Research (CIFAR) and the Natural Sciences and Engineering Research Council of Canada (NSERC).

\bibliography{refs}
\bibliographystyle{icml2018}

\clearpage
\appendix
\section{Understanding cross entropy loss in fair adversarial training}\label{sec:app-ce}

As established in the previous sections, we can view the purpose of the adversary's objective function as calculating a test discrepancy between $\mathcal{Z}_0$ and $\mathcal{Z}_1$ for a particular adversary $h$. 
Since the adversary is trying to maximize its objective, then a close-to-optimal adversary will have objective $L_{Adv}(h)$ close to the statistical distance between $\mathcal{Z}_0$ and $\mathcal{Z}_1$. 
Therefore, an optimal adversary can be thought of as regularizing our representations according to their statistical distance. 
It is essential for our model that the adversary is incentivized to reach as high a test discrepancy as possible, to fully penalize unfairness in the learned representations and in classifiers which may be learned from them.

However, this interpretation falls apart if we use (\ref{eq:bernoulli}) (equivalent to cross entropy loss) as the objective $L_{Adv}(h)$, since it does \textit{not} calculate the test discrepancy of a given adversary $h$.
Here we discuss the problems raised by dataset imbalance for a cross-entropy objective.

Firstly, whereas the test discrepancy is the sum of conditional expectations (one for each group), the standard cross entropy loss is an expectation over the entire dataset. 
This means that when the dataset is not balanced (i.e. $P(A = 0) \neq P(A = 1)$), the cross entropy objective will bias the adversary towards predicting the majority class correctly, at the expense of finding a larger test discrepancy. 
\begin{wraptable}{r}{3.75cm}
\begin{center}
\begin{small}
\begin{sc}
\begin{tabular}{ccc}
\toprule
  & $A = 0$  & $A = 1$ \\
\midrule
$Z = 0$     & 0.92 & 0.03  \\
$Z = 1$     & 0.03  & 0.02    \\
\bottomrule
\end{tabular}
\end{sc}
\end{small}
\end{center}
\caption{$p(Z, A)$}
\label{ce-example}
\end{wraptable} 
Consider the following toy example: a single-bit representation $Z$ is jointly distributed with sensitive attribute $A$ according to Table \ref{ce-example}. 
Consider the adversary $h$ that predicts $A$ according to $\hat A(Z) = T(h(Z))$ where $T(\cdot)$ is a hard threshold at $0.5$. 
Then if $h$ minimizes cross-entropy, then $h^*(0) = \frac{0.03}{0.95}$ and $h^*(1) = \frac{0.02}{0.05}$ which achieves $L(h) = -0.051$.
Thus every $Z$ is classified as $\hat A=0$ which yields test discrepancy $d_h(\mathcal{Z}_0, \mathcal{Z}_1) = 0$.
However, if we directly optimize the test discrepancy as we suggest, i.e., $L_{Adv}^{DP}(h) = d_h(\mathcal{Z}_0, \mathcal{Z}_1)$, $h^*(Z) = Z$, which yields $L_{Adv}^{DP}(h) = \E_{A = 0}[1 - h] + \E_{A = 1}[h]  - 1 =  \frac{0.92}{0.95} + \frac{0.02}{0.05} - 1 \approx 0.368$ (or vice versa). 
This shows that the cross-entropy adversarial objective will not, in the unbalanced case, optimize the test discrepency as well as the group-normalized $\ell_1$ objective.

\section{Training Details}\label{sec:training-details}

We used single-hidden-layer neural networks for each of our encoder, classifier and adversary, with 20 hidden units for the Health dataset and 8 hidden units for the Adult dataset. 
We also used a latent space of dimension 20 for Health and 8 for Adult. 
We train with $L_C$ and $L_{Adv}$ as absolute error, as discussed in Section \ref{theory}, as a more natural relaxation of the binary case for our theoretical results. 
Our networks used leaky rectified linear units and were trained with Adam \citep{kingma2014adam} with a learning rate of 0.001 and a minibatch size of 64, taking one step per minibatch for both the encoder-classifier and the discriminator. 
When training \textsc{ClassLearn} in Algorithm \ref{alg:laftr} from a learned representation we use a single hidden layer network with half the width of the representation layer, i.e., g.	
\textsc{ReprLearn} (i.e., LAFTR) was trained for a total of 1000 epochs, and \textsc{ClassLearn} was trained for at most 1000 epochs with early stopping if the training loss failed to reduce after 20 consecutive epochs.

To get the fairness-accuracy tradeoff curves in Figure \ref{results:pareto-fairness}, we sweep across a range of fairness coefficients $\gamma \in [0.1, 4]$. To evaluate, we use a validation procedure. For each encoder training run, model checkpoints were made every 50 epochs; $r$ classifiers are trained on each checkpoint (using $r$ different random seeds), and epoch with lowest median error $+ \Delta$ on validation set was chosen. We used $r = 7$.
Then $r$ more classifiers are trained on an unseen test set. The median statistics (taken across those $r$ random seeds) are displayed.

For the transfer learning expriment, we used $\gamma=1$ for models requiring a fair regularization coefficient.
\end{document}